\documentclass[lettersize,journal]{IEEEtran}
\usepackage{amsmath,amsfonts}
\usepackage{algorithmic}
\usepackage{algorithm}
\usepackage{array}
\usepackage{textcomp}
\usepackage{stfloats}
\usepackage{url}
\usepackage{verbatim}
\usepackage{graphicx}
\usepackage{cite}
\usepackage{indentfirst}
\usepackage{epstopdf}
\usepackage{float}
\usepackage{caption}
\usepackage{subcaption}
\usepackage{pgfplots}
\usepackage{booktabs}
\usepackage{tabularx}
\usepackage{xcolor}
\pgfplotsset{compat=1.17}
\usepackage{hyperref}
\usepackage{ragged2e} 
\usepackage{booktabs,makecell, multirow, tabularx}
\begin{document}
	
	\title{Dynamic Double Space Tower}
	\author{Sun Weikai~\IEEEmembership{Member,~IEEE}\thanks{For correspondence, please contact 15688532358@163.com\par This research is supported by Huawei Cloud Computing}, Song Shijie, and Wang Han}
	\markboth{Dynamic Double Space Tower}{}
	\maketitle
	
	\begin{abstract}
		The Visual Question Answering (VQA) task requires the simultaneous understanding of image content and question semantics. However, existing methods often have difficulty handling complex reasoning scenarios due to insufficient cross-modal interaction and capturing the entity spatial relationships in the image.\cite{huang2023adaptive}\cite{liu2021comparing}\cite{guibas2021adaptive}\cite{zhang2022vsa}We studied a brand-new approach to replace the attention mechanism in order to enhance the reasoning ability of the model and its understanding of spatial relationships.Specifically, we propose a dynamic bidirectional spatial tower, which is divided into four layers to observe the image according to the principle of human gestalt vision. This naturally provides a powerful structural prior for the spatial organization between entities, enabling the model to no longer blindly search for relationships between pixels but make judgments based on more meaningful perceptual units. Change from "seeing images" to "perceiving and organizing image content".A large number of experiments have shown that our module can be used in any other multimodal model and achieve advanced results, demonstrating its potential in spatial relationship processing.Meanwhile, the multimodal visual question-answering model July trained by our method has achieved state-of-the-art results with only 3B parameters, especially on the question-answering dataset of spatial relations.
	\end{abstract}
	
	
	\section{INTRODUCTION}
		\IEEEPARstart{V}{isual} Question Answering (VQA), as one of the core tasks of multimodal artificial intelligence, aims to generate accurate answers by combining visual content with natural language questions. In recent years, with the development of deep learning technology, VQA methods based on attention mechanisms and pre-trained models have significantly improved performance (such as the application of the Transformer architecture). However, the existing methods still have significant limitations in the following two aspects:\par
		
		1) Insufficient spatial relationship modeling: Although spatial attention mechanisms (such as spatial attention modules and multimodal fusion) are widely adopted, most models can only capture simple azimuth relationships (such as up and down, left and right), and have difficulty handling complex metric space reasoning (such as distance comparison, object size differences).\cite{zhou2024adapt}\cite{hu2024crd}\cite{sun2024generative}\cite{islam2023fast}\cite{liu2023cogan}\cite{wang2023image}For example, models such as SpatialVLM enhance spatial reasoning capabilities by synthesizing large-scale 3D data, but they rely on external knowledge bases and have high computational costs.\par
		
		2) The disconnection between interpretability and the reasoning process: Most existing methods enhance interpretability through posterior interpretation generation (such as the diffusion chain of thought model) or causal intervention, but there are still semantic deviations between the reasoning process and answer generation, especially in complex scenarios where it is difficult to gradually correlate problems with visual cues.\par
		
		3) Fine-grained spatial relationship modeling: The existing methods have limited ability to capture the dynamic spatial relationships (such as occlusion and relative motion) between entities in the image, resulting in complex problems (such as "Can object A be placed in container B?") The reasoning of) failed.\par
		
		The core of the Multimodal Visual Language (VQA) architecture is the concept of visual language modeling. These models are usually composed of three key steps. First of all, the single-modal visual architecture extracts meaningful information from the image. Typically, a visual encoder is a frozen visual Transformer (ViT), usually based on CLIP [17, 41]. Secondly, a projection module builds a bridge between vision and language, converting visual features into forms that language models can understand and process. This module is usually a simple linear layer or MLP [33, 34, 54], or a Transformer architecture based on cross-attention [6, 15, 31]. Finally, the projected visual information and text instructions (usually in the form of questions or prompts) are inserted into the large language model (LLM) to complete the task.\par
		
		Although significant progress has been made in visual Question answering (VQA) research, we have identified an interesting but often overlooked limitation in this type of architecture.The success of this type of model relying on the attention mechanism is due to its relatively successful fine-grained feature extraction in the image, but it ignores the complex spatial relationships in the image and the alignment and fusion of the text. This type of model in VQA answers questions simply by aligning pixels and text, which runs counter to the gestrar principle of human visual observation.\cite{li2023novel}\cite{wang2023rca}\cite{yang2023lpgan}\cite{jiang2023underwater} It has poor interpretability and is difficult to establish an effective system modeling, and has insufficient perception of space in the image. For example, when humans observe object A in object B, they rely on the continuity and closure of the objects, while the existing VQA models rely on the alignment of pixels and text in the image for observation. This undoubtedly has an impact on the robustness of the model and the understanding of spatial relationships. \par
		
		Therefore, we advocate improving the model's way of observing images based on the Gestalt principle of human visual observation to obtain better spatial relationship understanding ability and visual text fusion.\par
		
		Overall, our research provides a new paradigm for current spatial relationship modeling by introducing a dynamic two-stream spatial tower based on causal reasoning and a target detector based on vit, greatly enhancing the interpretability and robustness of the model. Moreover, it achieves the synergy of causal reasoning and statistical reasoning for the first time, filling the reasoning gap of "correlation → causality":\par
		
		1) We have created a dynamic two-stream space tower based on the principle of human visual gestalt to replace the attention mechanism. Help the model better learn the spatial relationships of objects in the image and achieve text-image beauty alignment. And a large number of experiments were conducted to prove its effectiveness.The accuracy and robustness of alignment have been improved, especially when dealing with complex or partially occluded objects.\par
		
		2) Through the counterfactual intervention module and the causal expert system, we have achieved the synergy of causal reasoning and statistical reasoning in a lightweight framework for the first time. Transform the statistical correlation drive of modal reasoning into a dual-drive mechanism of causality and statistics. The reasoning, alignment ability and speed of the model have been greatly improved when the amount of calculation is not large or even reduced.\par
		
		3) We have implemented a weighting strategy of continuous co-guidance of text and space to help the model better understand the problems and images when facing some highly difficult reasoning problems in VQA instead of rigidly observing through pixels. Truly enable the model to learn to think.\par
	
	\section{RELATED WORK}
		Visual Question Answering (VQA), due to its numerous application scenarios and development potential for future directions, such as medical image analysis, environmental understanding and decision-making, video and graphic understanding, and complex problem reasoning.It has attracted extensive attention from researchers in different fields.\cite{zhen2023cyclic}\cite{vela2023improving}\cite{zhu2023wdig}\cite{tian2020attentional}\cite{chen2022automatic}\cite{wang2024genartist}According to the application scenarios of the VQA model, we can classify it into three categories: 1) Medical analysis, 2) complex problem reasoning, and 3) multi-scenario interaction and general services. With the continuous development of deep learning, more and more researchers optimize VQA models through techniques such as loss functions, causal reasoning, attention mechanisms, and multimodal alignment fusion.\par
		
		\noindent A. Architecture\par
		
		Architecture improvement is a common and direct method for optimizing the VQA model. The main purpose is to enhance the model's ability to understand problems and images. Introducing new components is the main and common approach.\par
		
		Architecture improvement has two advantages: 1) It can adapt the architecture to different tasks and better facilitate task completion; 2) The improvement of the architecture can be carried out without increasing the computational load, and the improved architecture is expected to enhance performance. The disadvantages of the architecture improvement are as follows: 1) Although the improvement can be made without increasing the computational load, due to the existing thinking architecture of the VQA model, the computational load is still very large and the demand for computing resources is extremely high; 2) Limited generalization ability; 3) The interpretability of the model remains a problem. \par
		
		In our work, we use the dynamic two-stream spatial tower combined with the cause-statistical dual-driven mechanism, which greatly enhances the generalization ability and interpretability of the model. Moreover, with a relatively small amount of computation, its performance can be comparable to that of most open-source cutting-edge models such as llama and fusion, etc.\par
		
		\noindent B. Alignment fusion\par
		
		The alignment and fusion of modalities mainly focus on the high degree of fusion and alignment among multiple modalities (text-visual modalities), achieving the corresponding regions of semantics and images to enhance the model's ability to understand text and images. Multimodal fusion can enable the model to correlate different modalities.\cite{zhang2021joint}\cite{radford2021learning} In terms of fusion, it mainly includes methods such as pre-alignment fusion and post-alignment fusion. However, although the alignment and fusion between modalities can help the model understand the inputs of different modalities. However, there is a problem that multimodal fusion sometimes causes the model to have a wrong understanding of the visual modality (for example, the model may interpret the square text description as the corresponding square entities in the image modality), which is not friendly to some complex reasoning problems.\par
		Based on the principles of closure, continuity, proximity and similarity in human vision, we have solved this illusion problem that occurs during modal fusion. The model has achieved the most advanced effect in most spatial understanding problems.\par
		
		\noindent C. Reasoning\par
		
		The reasoning ability of the model is to enhance the model's ability to solve multi-step problems.\cite{li2023scaling}\cite{li2021align}\cite{wang2022image}\cite{li2022blip}\cite{li2023blip} At present, the reasoning aspect mainly includes three points: 1) Diffusion thinking chain, 2) parallel reasoning, and 3) causal reasoning. Reasoning can enable the model to understand the problems and images in VQA more deeply and give more accurate and comprehensive answers. However, the existing reasoning methods are difficult to bridge the gap between correlation and causality. We have achieved the synergy of causal reasoning and statistical reasoning in a lightweight framework for the first time through the cause-statistics dual-driven mechanism. It has significantly enhanced the reasoning ability and interpretability of the model.\par
		
	\section{METHODS}
		In this section, we describe the basic architecture of the model, the construction process of the dynamic two-stream space tower, the implementation of the cause-statistics dual-driven mechanism, and the implementation of the text space-guided encoding mechanism..\par
		
		\noindent A. Dynamic double space tower\par
		
		Our dynamic two-stream space tower is constructed based on the proximity, continuity, closure and similarity of the gestalt.\par
		
		Firstly, regarding proximity, in the Gestalt principle, proximity refers to the fact that elements that are geographically close are easily regarded as a whole. We introduce a factor related to spatial distance. For image patches or region proposals, calculate the spatial distance between them (for example, Euclidean distance, Manhattan distance). Integrate this distance information into the calculation of the space tower. In the calculation, algorithms such as SLIC are used to generate superpixels, and each superpixel participates in the calculation of the spatial tower as a basic unit. The image regions are regarded as the nodes of the graph, and edges are established between the spatially adjacent regions. GNN can propagate information on the graph structure and naturally integrate proximity. Attention can be calculated at the node (region) level and is influenced by the graph structure (the connection of edges).\par
		
		\begin{figure}[H]
			\centering
			\includegraphics[width=8cm,height=6cm]{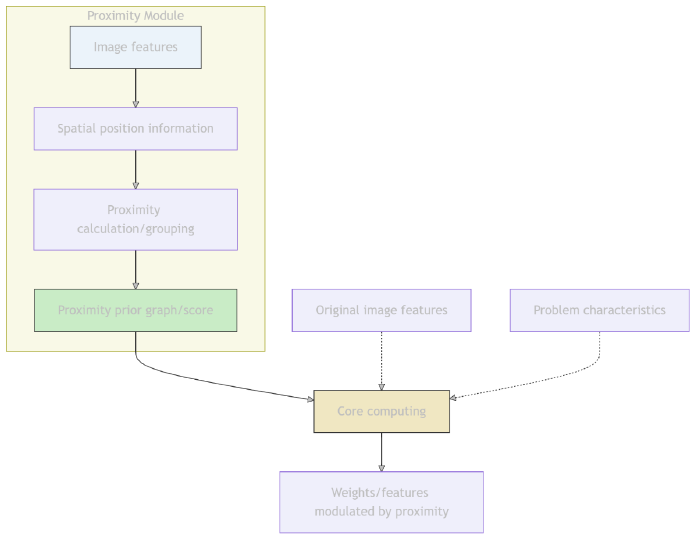}
			\caption{Proximity architecture}
		\end{figure}
		
		Secondly, regarding similarity, it refers to the fact that elements with similar attributes will be automatically grouped into the same group or whole by the brain, even if they are not spatially close. This similarity can be reflected in multiple visual dimensions. The brain simplifies information processing and enhances perceptual efficiency by capturing these commonalities. Based on this, we make the spatial tower pay more attention to those areas that are relevant to the query area or context in terms of visual features (color, texture, shape, semantics, etc.) rather than other irrelevant backgrounds and other semantically irrelevant areas. The specific implementation is to extract the feature vectors of different regions of the image (for example, deep features from CNN). Calculate the similarity between these feature vectors (such as cosine similarity, reciprocal of L2 distance). Cluster the image regions based on features. The spatial tower can be biased towards the regions within the same cluster or focus on those clusters that are similar to the object features described in the problem.\par
		
		\begin{figure}[H]
			\centering
			\includegraphics[width=8cm,height=6cm]{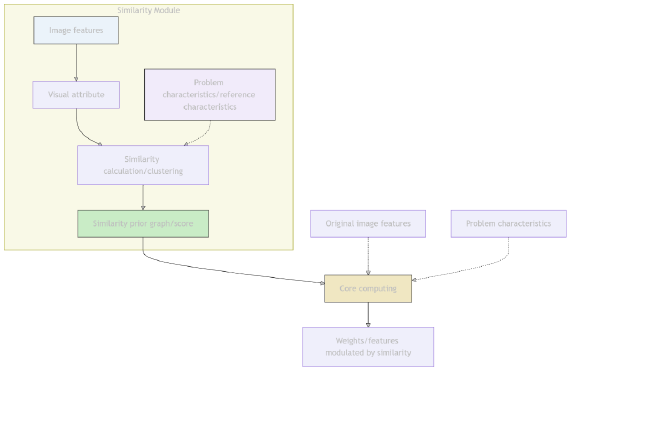}
			\caption{Similarity architecture}
		\end{figure}
		
		Thirdly, for closure, it refers to the fact that the brain will automatically fill in incomplete graphics to make them a complete whole. We assume that when the model recognizes certain parts of an object, it should be able to infer the outline or range of the complete object and direct attention to this "closed" area, even if some parts are occluded or unclear. We combine classic computer vision technology or edge detection modules of deep learning. If non-closed but nearly closed edges are detected, connect them through the Virtual bridge and guide the spatial tower to this inferred closed area. We also designed a loss function to penalize the closed areas that the spatial tower failed to recognize, helping the spatial tower better allocate perception to "cover" the entire object.\par
		
		\begin{figure}[H]
			\centering
			\includegraphics[width=8cm,height=6cm]{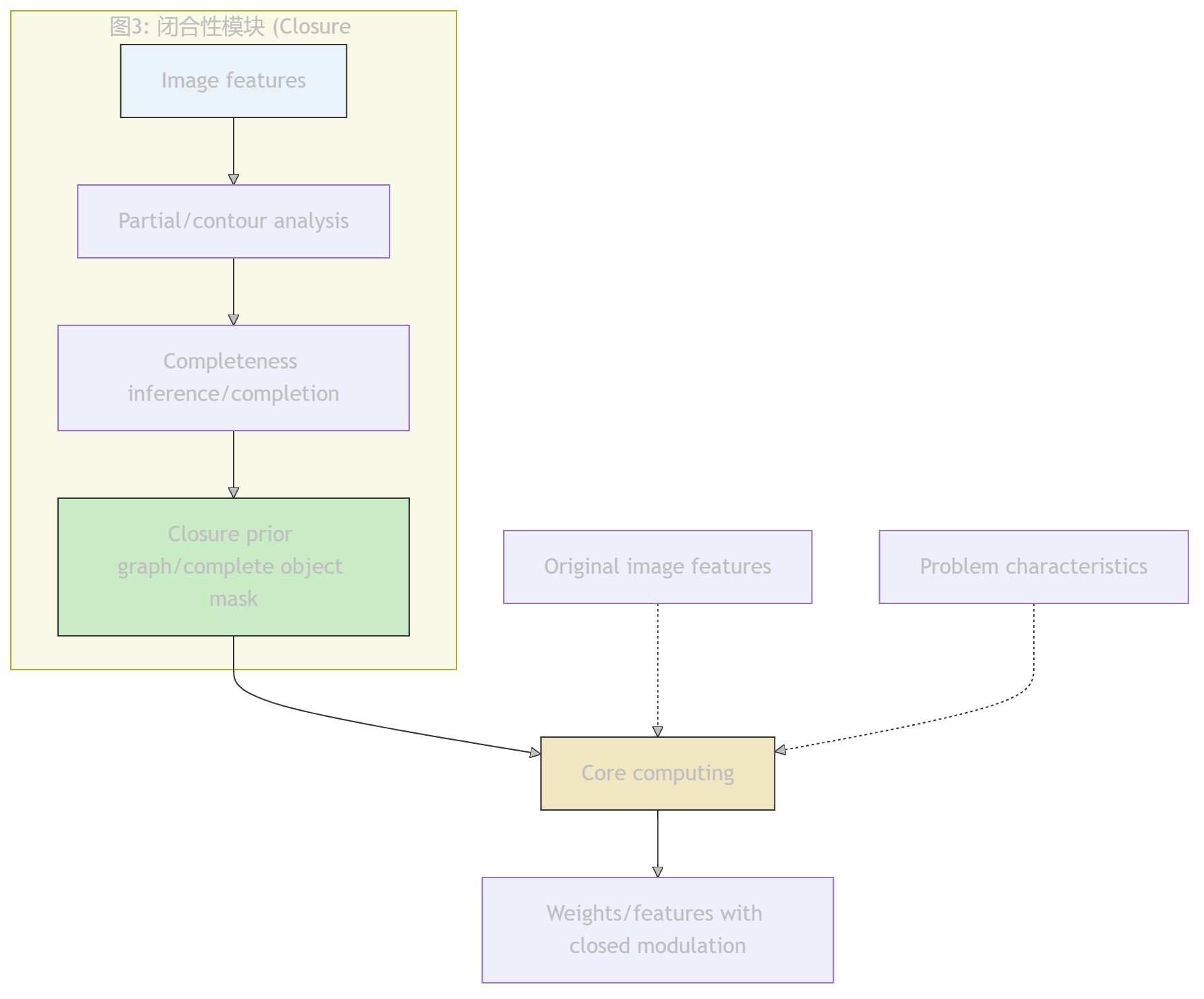}
			\caption{Closure architecture}
		\end{figure}
		
		Fourth, regarding continuity, it refers to the visual tendency to perceive continuous lines or shapes rather than broken fragments. Let's make the model for something like "along..." "Through..." When it comes to the relationship of paths or directions, we guide the continuous module to gradually move in the direction of the path through the weights guided by the text space. "Remember" the spatial positional relationships in the image through paths and assign different weights. To avoid the problem of path "loss" when guiding the path, we separate the background and entities by calling the closed-integration lightweight framework.\par
		
		\begin{figure}[H]
			\centering
			\includegraphics[width=8cm,height=6cm]{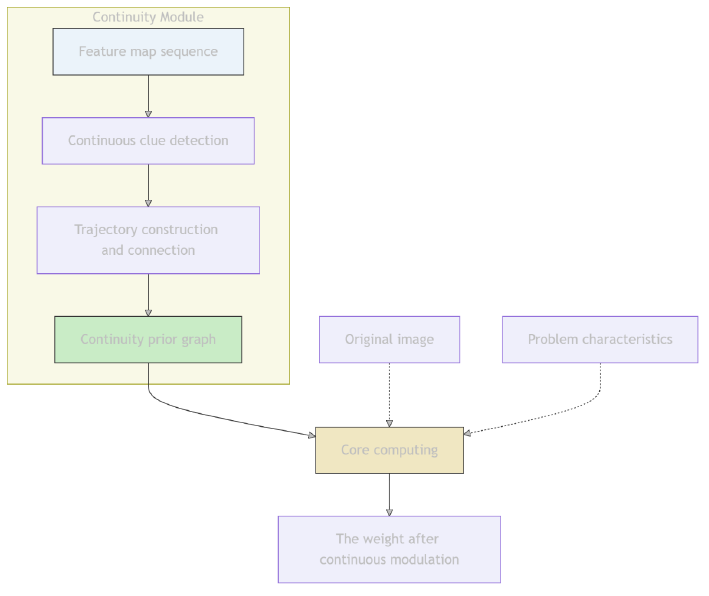}
			\caption{Continuty architecture}
		\end{figure}
		
		\noindent B. Model architecture\par
		
		Our model consists of three parts. The first part is the encoding part, the second part is the embedding and fusion, and the third part is the decoding and generation part.\par
		
		For the encoding part, we choose to improve on the basis of clip and integrate the dynamic dual-stream space tower. The dynamic dual-stream spatial tower influences the extraction of image features and text feature weights through four layers of modules. The correspondence between text and image entities (such as "fire source" corresponding to "heat") is achieved only through training without going through an external knowledge base. Meanwhile, while maintaining the computational complexity, it also greatly enhances the model's perception and thinking ability for complex spatial problems, achieving a significant improvement in spatial modeling.\par
		
		We adopt a double-layer fusion approach to fuse the modalities. Firstly, the modalities are mapped into the joint embedding space through the dynamic spatial tower and the cross-causal weights guided by the text to achieve the initial fusion of the text and the image. Then, the dynamic interaction fender and the decoding layer of causal intervention are deeply fused for the second time, and the cause-statistical dual-driven mechanism is utilized to achieve further alignment and fusion between modalities.\par
		
		In the decoding generation section, we utilize a dynamic two-stream space tower and causal score weighting to influence the weights of the decoder and the hybrid expert system. Achieve causal consistency decoding and promote spatial alignment. To ensure computational efficiency, we adopt sparsity processing to achieve lightweight. The causal expert system adopts parameter sharing (for example, "causal reasoning expert" and "statistical reasoning Expert" share the underlying network, with only the top-level output layer being independent). We also visualized the causal intervention heat map (such as showing which answer tokens' probabilities changed significantly after "deleting the fire source"), and generated the causal chain explanations of the reasoning process (such as "The water boiled because there was a fire source heating the kettle").\par
		
		\noindent C. Dual-driven by causality and statistics\par
		
		By integrating lightweight causal modeling and counterfactual intervention mechanisms and adding cross-causal modeling guided by the text space, we can explicitly identify and utilize causal relationships between modalities, thereby overcoming the problem of false association and better answering questions involving "why" or "if..." Questions about deep causal relationships such as "What will happen?".\par
		The decoder is designed to adapt to both specific task types and potential causal intentions simultaneously to ensure the accuracy and causal consistency of the generated answers. To take into account the computational efficiency, we only insert causal intervention layers in the last two layers of the decoder. Through intervention operations (such as blocking the word "because" in the text), the model tests the robustness of the generated results. If the results change significantly, the model will strengthen the corresponding causal dependency weights to avoid dependency surface associations. This operation increases the amount of calculation by less than 5\%. Achieve end-to-end precise mapping of "question word → visual causal entity → answer token" (for example, "Why is the kettle steaming?") → Locate the "Fire Source" patch → Activate the "Heating → Evaporation" causality Expert.
		
		\noindent D. Encoding guided by text space\par
		
		Causal intent token: Add special tags [CAUSE] and [EFFECT] at the beginning and end of the question text to explicitly mark the causal reasoning intent. For example, "Why is there steam coming from the kettle?" Reconstructed as [CAUSE] Why is the kettle steaming? [EFFECT]. Causal trigger word mask: Use spaCy/NLTK to detect causal trigger words (such as "because", "cause", "so"), and generate a binary mask vector C\_mask (the position of the trigger word is 1, and the rest is 0) to ensure the integrity of the trigger word is retained during word segmentation.\par
		The position encoding of each word is composed of the original position vector and the causal role vector. Determine the causal role (cause/effect/irrelevant) of words through dependency syntactic analysis or pre-trained dictionaries (such as CauseNet). For example, "fire source" is assigned the cause role vector E\_cause, "boiling" corresponds to the result role vector E\_effect, and irrelevant words use the zero vector.\par
		In the fine-tuning stage, randomly MASK or replace the causal trigger words in the text (for example, change "Because A, so B" to "[MASK] A, [MASK] B"), requiring the model to predict the causal role (cause/effect/irrelevant) of the masked words, and drive the model to learn causal dependencies through the cross-entropy loss L\_causal-cls.Meanwhile, in the Dropout operation, we forcibly retain the embedding of causal trigger words to prevent the model from ignoring explicit causal signals.\par
	
	\section{EXPERIMENTAL}
		In the experimental work, we used a variety of benchmark methods to verify our model and method. The results show the effectiveness and advancement of our method.\par
		
		Training strategy: We introduced a three-stage training framework for training to ensure the deep integration and alignment between visual and language modalities.\par
		
		Phase One: Basic semantic-image alignment, using large-scale image-text pair datasets and pre-training encoders to establish precise semantic-pixel alignment.\par
		
		Phase Two: Unsupervised Causal representation learning, using large-scale unlabeled graphic and text data, and extracting causal invariant features through causal-AE and ICA contrastive learning.\par
		
		Phase Three: Supervised fine-tuning, trained on datasets such as VQA and ScienceQA, to activate the causal intervention module.\par
		
		In the first phase, we used a large number of image-text datasets, including the COCO dataset, PixelProse dataset, Llama558k dataset, and we also used some publicly available datasets.\par
		
		In the second phase, we used multiple domain-specific datasets, such as those in mathematics, medicine, educational Q\&A, spatial reasoning, and visual reasoning, to further enrich the dataset. In Phase 2, we focused on training the alignment fusion of visual and textual data, particularly the dynamic bidirectional spatial tower. Training the bidirectional spatial tower enabled the model to achieve better spatial perception and reasoning capabilities.\par
		
		Implementation details of the dynamic dual-stream space tower: We use the CNN network as the benchmark network, and build our space tower on this basis to replace the traditional attention mechanism. CNNs disseminate information on the graph structure, integrating proximity naturally. Attention can be computed at the node (region) level and is affected by the graph structure (the connection of edges). We use the design loss function to optimize the space tower. We use the vit-based visual detection model to obtain the mask of the complete object, calculate the metric IoU between the space tower prior map and the complete object mask, and add it to the loss function to prompt the space tower to generate a priori map to "cover" the entire object object.\par
		
		\noindent We tested and evaluated our model using benchmark datasets in three areas. They are multimodality, visual reasoning, and spatial relationship perception:\par
		
		Multimodal datasets: MMBench, POPE, and LLAVA.\par
		
		Visual reasoning: CLEVR dataset, VSRE dataset, NuminaMath-CoT dataset.\par
		
		Spatial relationship perception: CV-Bench dataset, VSIBENCH dataset.\par
		
		\begin{table*}[htbp]
			\caption{Comparisons of other parameter models on 8 benchmarks.}
			\label{tab:model_comparison}
			\subcaption{Results on general multimodal benchmarks.}
			\centering
			\resizebox{\textwidth}{!}{
				\begin{tabular}{l cccccccc}
					\toprule
					\textbf{Model} & \textbf{path-vqa} & \textbf{vqa-rad} & \textbf{plot-qa} & \textbf{POPE} & \textbf{CV-Bench} & \textbf{VSI-Bench} & \textbf{CLEVR} & \textbf{MMBench}\\[5pt]
					\midrule
					Qwen2.5VL 3B & 79.1 & 78.1 & - & 85.9 & 61.4 & 74.0 & 46.6 & 56.3 \\[5pt]
					InternVL2 4B & 78.5 & 73.9 & - & 84.6 & 50.5 & 73.2 & 42.4 & 53.9 \\[5pt]
					DeepSeek-VL2-Tiny & 74.6 & 72.1 & - & - & 52.5 & 72.3 & 39.6 & 45.9 \\[5pt]
					MM1.5 3B & - & - & - & 88.1 & 41.0 & 72.4 & - & - \\[5pt]
					Phi 3.5-Vision & 75.5 & 64.2 & 58.2 & 82.2 & 46.5 & 69.9 & 53.3 & 49.0 \\[5pt]
					Florence-VL 3B & 71.6 & 60.8 & 59.1 & 88.3 & 51.0 & 70.6 & 58.1 & 44.9 \\[5pt]
					\midrule
					\textbf{July(Ours)} & \textbf{79.6} & \textbf{70.9} & \textbf{66.1} & \textbf{88,7} & \textbf{64.6} & \textbf{79.8} & \textbf{65.1} & \textbf{55.6} \\[5pt]
					\bottomrule
				\end{tabular}
			}
		\end{table*}
		
		As shown in Figure 1, our model demonstrates advanced performance in most benchmark tests. Particularly, in benchmark tests that require reasoning, our model outperforms models with the same parameter size in related issues such as mathematics, logical thinking, and spatial relationship perception reasoning. It is also noteworthy that our model only requires 640 tokens to compare with the current state-of-the-art model. Additionally, our model achieves the best results on some spatial perception datasets, emphasizing the effectiveness and advanced nature of our model.\par
		
		\noindent \textbf{Ablation experiments}\par
		
		In this section, we compare our model with the llavanext model to validate the effectiveness of the components in our model. We conducted a series of ablation studies. Here, we only explain the effect of the dynamic dual-stream spatial tower module. We selected a llava model with the same parameter count, removed the attention mechanism, and embedded our dynamic dual-stream spatial tower for testing and evaluation. The results show that our model with the embedded dynamic dual-stream spatial tower achieved a 12 percentage point higher accuracy on VISTBENCH compared to the model using traditional attention mechanisms, proving the advanced performance of our innovative dynamic dual-stream spatial tower.\par
		
		Our text space-guided cross-causal weights also play a significant role. We conducted a comparison ablation study using traditional attention weights and our text space-guided cross-causal weights. The results showed that in most datasets, the accuracy of our weight allocation strategy was higher than that of traditional attention weights. This demonstrates that our weight allocation strategy can better adapt to complex reasoning problems in visual question answering, providing the model with better robustness.\par
	
	\section{CONCLUSIONS AND FUTURE WORK}
		This study proposes an innovative dynamic bidirectional spatial tower architecture, aiming to overcome the current challenges of visual question answering by imitating the principle of human visual observation of the world and combining existing deep learning techniques. The experimental results show that our model has achieved advanced effects in VQA reasoning, especially in the understanding and reasoning of the spatial relationships of entities in images.\cite{huang2023language}\cite{raffel2020exploring}\cite{kim2021vilt}\cite{li2023vision}\cite{bao2022vlmo} It outperformed cutting-edge models such as llama and fusion in multiple benchmark tests.\par
		
		In conclusion, DDST not only provides a perspective for understanding the spatial relationship of images and reasoning ability, but also lays a solid foundation for subsequent research and development. However, we admit that in practical applications, problems such as long training time and high demand for hardware resources still need to be faced. This will be one of the directions that need to be addressed in future work.\par
		
		With the continuous advancement of artificial intelligence technology, it is expected that the field of VQA inference will witness more exciting development opportunities. For the DDST architecture, we look forward to the following potential research directions:\par
		
		1): Accelerate training and inference: Explore more efficient algorithms or hardware solutions to shorten the training cycle and reduce inference latency, making real-time VQA inference generation possible.\par
		
		2): Cross-modal learning: Further enhance the interaction among text, images, and other modal information to achieve a more natural and smooth human-computer collaboration experience.\par
		
		3): Adaptive learning ability: Develop intelligent systems with self-evaluation and adjustment functions so that the model can automatically optimize its own parameter configuration according to different tasks, improving generalization ability and robustness.\par
		
		4): Ethical Considerations and Social Impact: A deep exploration of the social responsibility of AI-generated content to ensure that technological development is ethical and beneficial to human society. In conclusion, the MSPG-SEN architecture represents an important milestone in image generation technology, and its future development potential depends on the spirit of continuous exploration and technological innovation of researchers. We look forward to every breakthrough in this field, bringing people an unprecedented visual feast and injecting new vitality into all industries.\par
	
	\section*{REFERENCES}\label{sec7}
	\bibliographystyle{plain}
	\bibliography{thebibliography.bib}
	\begin{IEEEbiography}[{\includegraphics[width=1in, height=1.25in, clip, keepaspectratio]{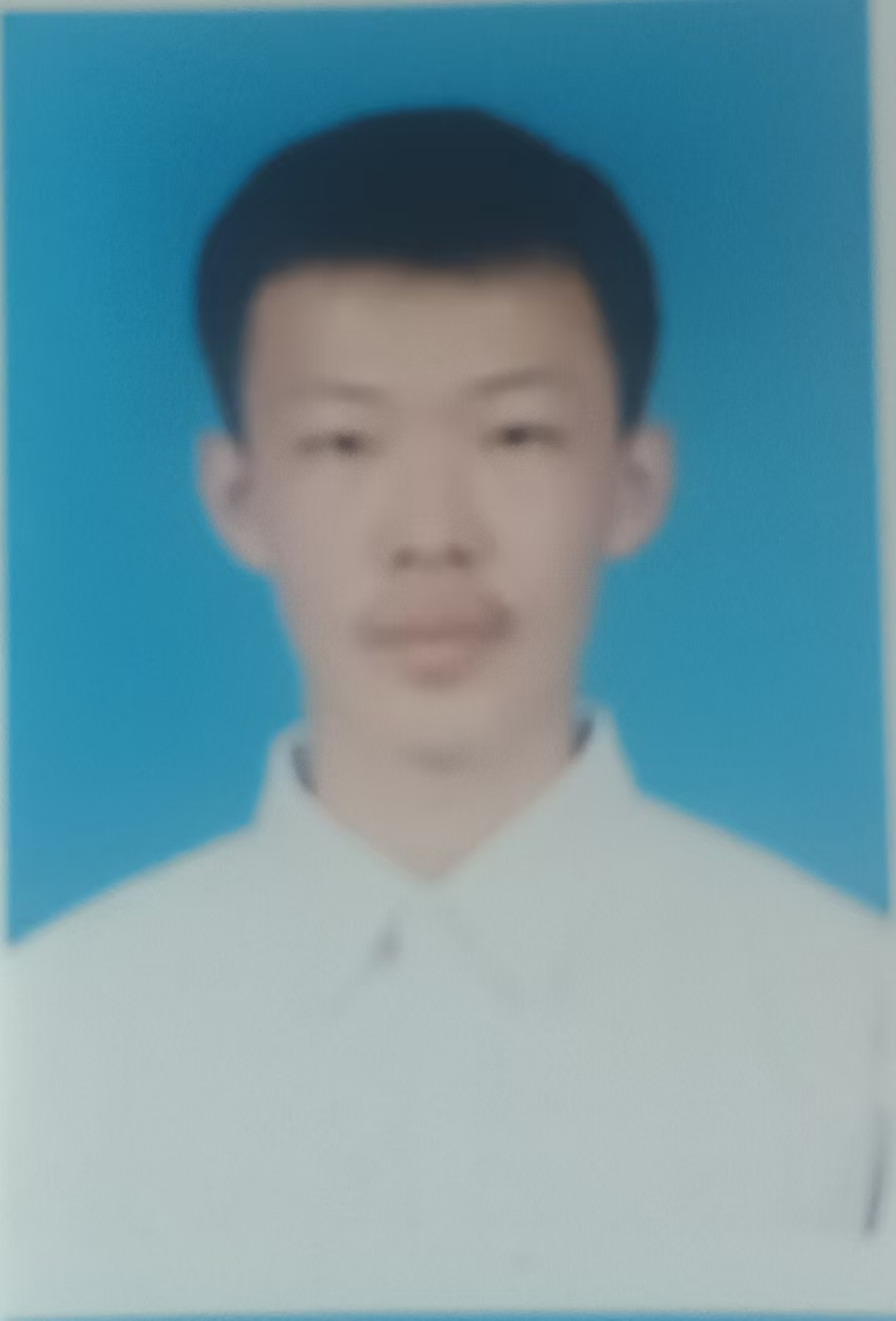}}]{Sun Weikai}
		He is currently studying for a Bachelor's degree in Computer Science and Technology at Qilu Normal University. \par
		His research focuses on multimodality, deep learning, operator development, and algorithm optimization. \par
		15688532358@163.com
	\end{IEEEbiography}
	\begin{IEEEbiography}[{\includegraphics[width=1in, height=1.25in, clip, keepaspectratio]{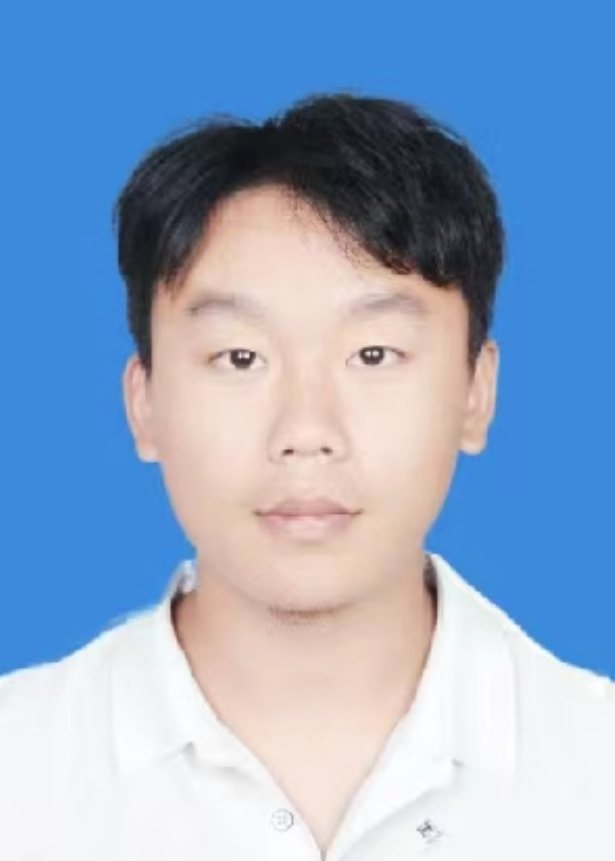}}]{Song Shijie}
		He is currently studying for a Bachelor's degree in Computer Science and Technology at Qilu Normal University. \par
		His research focuses on deep learning, multimodality, and image generation. \par
		1076271275@qq.com
	\end{IEEEbiography}
	\begin{IEEEbiography}[{\includegraphics[width=1in, height=1.25in, clip, keepaspectratio]{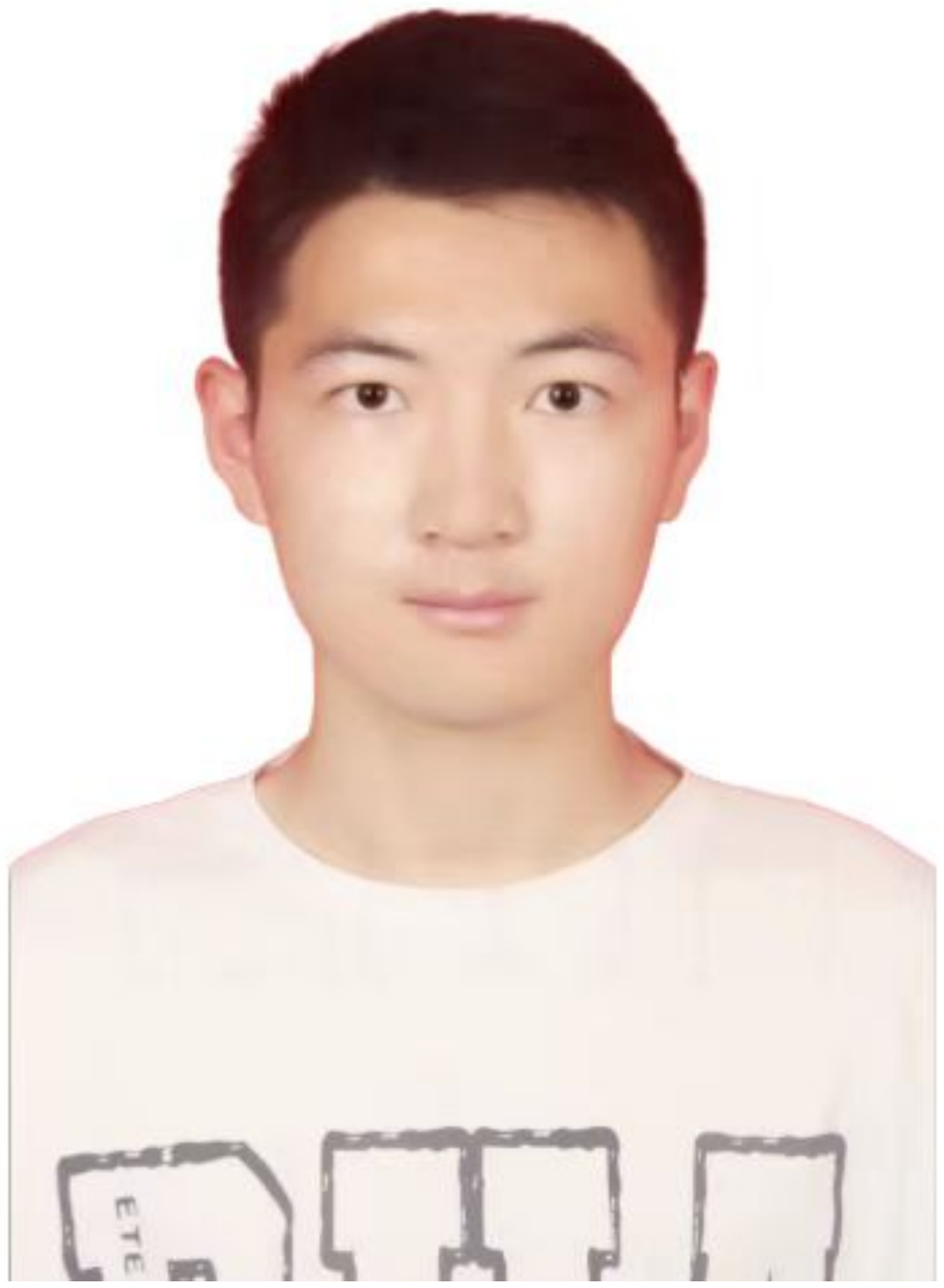}}]{Wang Han}
		He is currently studying for a Bachelor's degree in Computer Science and Technology at Qilu Normal University. \par
		His research focuses on deep learning, emotional computing, Embedded and Autonomous driving. \par
		2051059438@qq.com
	\end{IEEEbiography}
	
\end{document}